\documentclass{article}

\usepackage[preprint, nonatbib]{neurips_2024}

\usepackage[utf8]{inputenc} % allow utf-8 input
\usepackage[T1]{fontenc}    % use 8-bit T1 fonts
\usepackage{hyperref}       % hyperlinks
\usepackage{url}            % simple URL typesetting
\usepackage{booktabs}       % professional-quality tables
\usepackage{amsfonts}       % blackboard math symbols
\usepackage{nicefrac}       % compact symbols for 1/2, etc.
\usepackage{microtype}      % microtypography
\usepackage{xcolor}         % colors

\usepackage{amsmath, amsthm, amssymb}
\usepackage{dsfont}
\usepackage[ruled,vlined]{algorithm2e}
\definecolor{commentcolor}{RGB}{110,154,155}   % define comment color
\newcommand{\PyComment}[1]{\ttfamily\textcolor{commentcolor}{\# #1}}  % add a "#" before the input text "#1"
\newcommand{\PyCode}[1]{\ttfamily\textcolor{black}{#1}} % \ttfamily is the code font

\usepackage{verbatim}
\usepackage{graphicx}

\title{Maximum Manifold Capacity Representations in State~Representation~Learning}

\author{Li Meng$^1$, Morten Goodwin$^{2}$, Anis Yazidi$^{3,4}$, Paal Engelstad$^1$\\
$^1$University of Oslo\\
$^2$Centre for Artificial Intelligence Research, University of Agder\\
$^3$Oslo Metropolitan University\\
$^4$Norwegian University of Science and Technology}

\begin{document}
\maketitle

\begin{abstract}

The expanding research on manifold-based self-supervised learning (SSL) builds on the manifold hypothesis, which suggests that the inherent complexity of high-dimensional data can be unraveled through lower-dimensional manifold embeddings. Capitalizing on this, DeepInfomax with an unbalanced atlas (DIM-UA) has emerged as a powerful tool and yielded impressive results for state representations in reinforcement learning. Meanwhile, Maximum Manifold Capacity Representation (MMCR) presents a new frontier for SSL by optimizing class separability via manifold compression. However, MMCR demands extensive input views, resulting in significant computational costs and protracted pre-training durations. Bridging this gap, we present an innovative integration of MMCR into existing SSL methods, incorporating a discerning regularization strategy that enhances the lower bound of mutual information. We also propose a novel state representation learning method extending DIM-UA, embedding a nuclear norm loss to enforce manifold consistency robustly. On experimentation with the Atari Annotated RAM Interface, our method DIM-C$^+$ improves DIM-UA significantly with the same number of target encoding dimensions. The mean F1 score averaged over categories is 78\% compared to 75\% of DIM-UA. There are also compelling gains when implementing SimCLR and Barlow Twins. This supports our SSL innovation as a paradigm shift, enabling more nuanced high-dimensional data representations.

\end{abstract}

\section{Introduction}

State representation learning (SRL) is a particular case of representation learning in which the features to learn evolve through time, and are influenced by actions of an agent to interact with the environment \cite{lesort2018state}. Those learned state representations are useful for downstream tasks in the field of reinforcement learning (RL) and robotics \cite{jonschkowski2015learning}.

DeepInfomax with an unbalanced atlas (DIM-UA) \cite{meng2023state} is an SRL method that learns state representations in the unbalanced atlas (UA) of a manifold. Membership probabilities are assigned to each chart of an atlas in DIM-UA, in which UA means that the distribution of probabilities is designated to move away from uniformity.

Multi-view self-supervised learning (MVSSL) is a sub-field within the broader domain of self-supervised learning (SSL), which is powerful to learn feature representations from multiple transformations (views) of unsupervised data. In contrast to generative modeling approaches like variational autoencoders (VAEs) \cite{kingma2013auto} and adversarial autoencoders (AAEs) \cite{makhzani2015adversarial}, MVSSL typically leverages metrics that resonate with principles of information theory \cite{bachman2019learning}.

Contrastive MVSSL methods include well-known names such as Contrastive Predictive Coding (CPC) \cite{oord2018representation}, SimCLR \cite{chen2020simple}, and Momentum Contrast (MoCo) \cite{he2020momentum, chen2020improved}. Moreover, there are also clustering MVSSL methods \cite{bojanowski2017unsupervised, caron2018deep, caron2020unsupervised}, momentum-based MVSSL methods \cite{chen2021exploring, grill2020bootstrap}, as well as MVSSL methods reducing redundancy \cite{zbontar2021barlow, zhu2022tico}. As an outlier to the above families, Maximum Manifold Capacity Representation (MMCR) is capable of achieving state-of-the-art results based on the manifold capacity theory \cite{yerxa2024learning}.

In this study, we explore the integration of manifold capacity theory with existing MVSSL techniques. We present a novel SRL approach, named as DIM-C$^+$, grounded in DIM-UA, of which the manifold representations are refined by incorporating manifold capacity insights.

We summarize our contributions as follows:
\begin{itemize}
    \item We achieve the state-of-the-art results in SRL surpassing DIM-UA \cite{meng2023state} through the infusion of manifold capacity theory. Our method is also compared to other SRL paradigms on the Atari Annotated RAM Interface (AtariARI) \cite{anand2019unsupervised}. The empirical evidence demonstrates that our method delivers superior F1 and accuracy scores.
    \item We further demonstrate the potential to augment current MVSSL methodologies using the principles of manifold capacity theory. Notably, our approach requires only two views, thereby achieving improvements while introducing a modest computational demand.
    \item Our method is capable of achieving better results than DIM-UA without the need of a membership probability distribution that is moved away from uniformity. This yields a simplified yet more effective paradigm in SSL.
\end{itemize}

\section{Related Work}
\paragraph{Multi-view self-supervised learning} Contrastive learning has been a prominent approach in MVSSL, where representations are learned by distinguishing between positive and negative pairs of data points. Among those approaches, CPC \cite{oord2018representation} extracts representations by predicting the future in latent space using autoregressive models. MoCo \cite{he2020momentum} introduces a dynamic dictionary with a queue and a moving-averaged encoder, enabling consistent feature learning over shifting negatives. SimCLR \cite{chen2020simple} provides a straightforward yet extremely effective learning framework, demonstrating that sophisticated architectures are less crucial when equipped with proper data augmentations.

Recognizing the challenges associated with using negative pairs, momentum-based MVSSL methods have been introduced. BYOL \cite{grill2020bootstrap} features a momentum encoder, eliminating the need for negative pairings entirely. SimSiam \cite{chen2021exploring} further provides the insight that a separate momentum-updated network is not needed, and a stop-gradient operation is sufficient to prevent collapse.

Alternatively, Noise As Targets (NAT) \cite{bojanowski2017unsupervised} was introduced to constrain the latent features. DeepCluster \cite{caron2018deep} finds the natural grouping tendencies of data with k-means \cite{macqueen1967some} and iteratively groups the features. Swapping Assignments between Views (SwAV)\cite{caron2020unsupervised} uses a swapped prediction task where it matches the cluster assignments across views. The Barlow Twins (BT) method \cite{zbontar2021barlow} contributes the intriguing idea of reducing redundancy between the outputs of two identical networks that process distorted variants of the same input. Transformation Invariance and Covariance Contrast (TiCo) \cite{zhu2022tico} integrates the contrastive and redundancy-reduction methods and can be seen as a combined modification of MoCo and Barlow Twins.

\paragraph{Manifold representations} AAEs were used to learn an atlas capable of recovering the topology or homotopy type of a space from input in \cite{korman2018autoencoding}. Building upon this foundation, manifold representations through MSimCLR \cite{korman2021atlas, korman2021self} extend this paradigm, of which the output incorporates chart embeddings along with membership probabilities. The gap between linear and nonlinear manifold approaches can be bridged by a combination of autoencoders and BT \cite{kadeethum2022reduced}. The encoder-decoder pairs of a generative model can also utilize a Riemannian gradient descent algorithm to learn chart representations of a manifold \cite{alberti2023manifold}.

\paragraph{State representation learning} Unsupervised SRL methodologies have been employed both explicitly and implicitly across a spectrum of studies \cite{caron2019unsupervised, laskin2020curl, stooke2021decoupling, eysenbach2022contrastive, meng2023unsupervised}. The Spatiotemporal DeepInfomax (ST-DIM) approach \cite{anand2019unsupervised} builds on DeepInfoMax (DIM)\cite{hjelm2018learning} and maximizes the mutual information between spatial and temporal features, achieving notable outcomes on the Atari Annotated RAM Interface (AtairiARI) benchmark. DIM-UA \cite{meng2023state} further integrates manifold representations into SRL by equipping the atlas of the manifold with an unbalanced objective, demonstrating the state-of-the-art performance on AtairiARI.

\section{Method}
\subsection{Čech nerve}

\begin{figure}[t]
    \centering
    \includegraphics[width=0.7\linewidth]{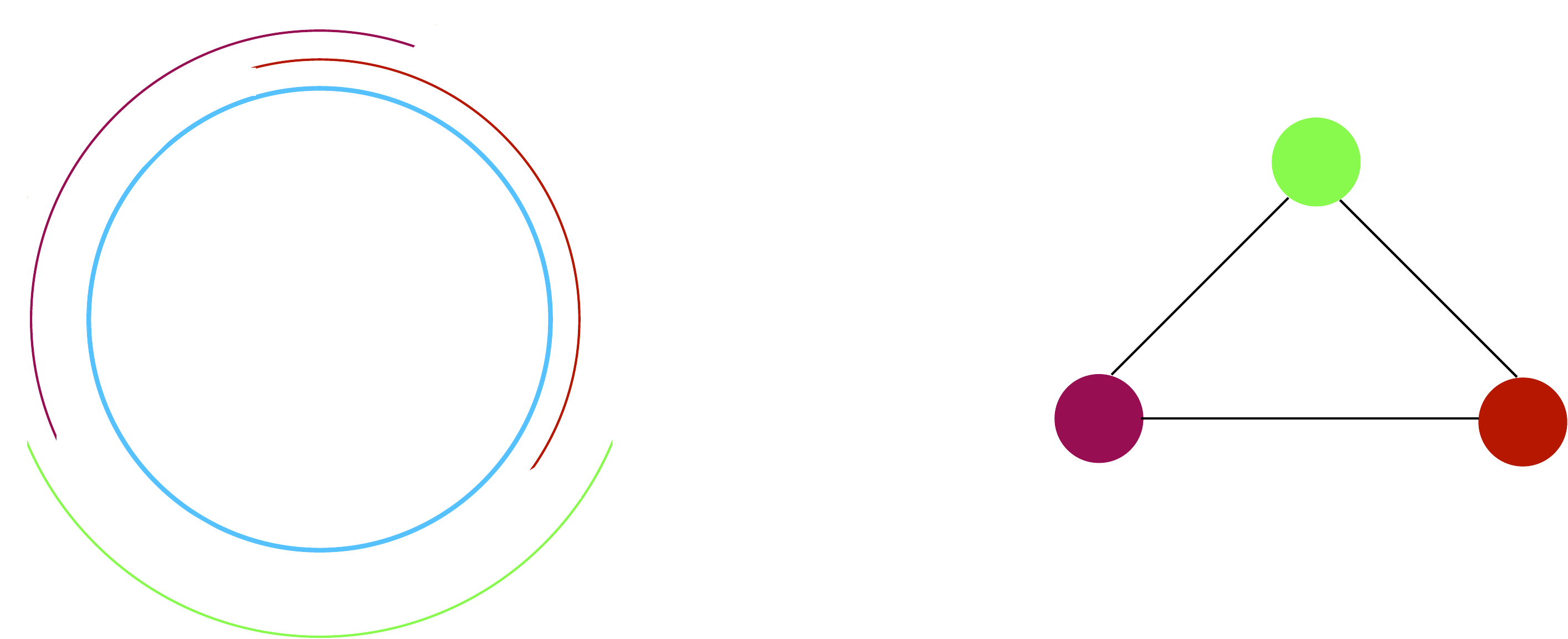}
    \caption{Presented on the left is an atlas of the circle, while on the right side is the Čech nerve of the circle. Each vertex of the graph indicates an individual chart of the atlas. Edges are drawn between pairs of vertices to indicate that their respective charts intersect with a non-empty region of overlap.}
\label{fig:cech}
\end{figure}

The unsupervised manifold learning paradigm adopted in this work was initiated for the purpose of recovering the topology or homotopy of a space from sample points \cite{pitelis2013learning, korman2018autoencoding}. In the special case where the intersection of any collection of charts is contractible, the homotopy type of the manifold can be recovered through a combinatorial object: the Čech nerve. An illustration of the Čech nerve for a circle $S^1 \subset \mathbb{R}^2$ is shown in Fig. \ref{fig:cech}. In the case of submanifolds embedded in $\mathbb{R}^n$, each chart of an atlas emerges as the restriction of a linear mapping $ \mathbb {R}^n \rightarrow \mathbb{R}^d $.

\subsection{Encoding space}
For a distribution of input data $\mathcal{X}$ and its latent space $\mathcal{Z}$ with an embedding function, $z$ the embedding of input from input space $\mathcal{X}$: $\mathcal{X} \rightarrow \mathcal{Z}$, we have $d$ dimensions for each output $o$ of $n$ charts, and we use $\mathcal{N}$ to denote $\{1, 2, ..., n\}$. A manifold  \cite{korman2021atlas, korman2021self, meng2023state} can be encoded by two components: 
\begin{enumerate}
    \item $\phi_i$: $\mathcal{Z} \rightarrow \mathbb{R}^d$, for $i \in \mathcal{N}$, is the inverse mapping of a coordinate map: $\mathbb{R}^d \rightarrow \mathcal{Z}$. For random variables $I, Z$ on $\mathcal{N}, \mathcal{Z}$, we have $p(o \; | \; I=i,Z=z) = \mathds{1}_{\phi_i(z)}$, where $\mathds{1}_{\phi_i(z)}$ indicates a distribution supported at a single point $\phi_i(z)$.

    \item $q: \mathcal{Z} \rightarrow [0, 1]^n$ is the chart membership function. For each $i \in \mathcal{N}$, $q_i(z) = p(I=i \; | \; Z=z)$.
\end{enumerate}

The posterior can be calculated by Eq. \ref{eq:posterior}, and therefore, the prior of the encoding space can be obtained by Eq. \ref{eq:prior}.

\begin{equation}
p(o,i\; | \;z) =p(o\; | \;i,z)p(i\; | \;z)=q_i(z)\mathds{1}_{\phi_i(z)}
\label{eq:posterior}
\end{equation}

\begin{equation}
p(o,i) =\mathbb{E}_z q_i(z)\mathds{1}_{\phi_i(z)}
\label{eq:prior}
\end{equation}

At inference time, we only choose $i = \mathrm{argmax}_j\; q_j(z)$ and use a one-hot encoding of $q(z)$.

\subsection{Unbalanced atlas}
UA adopts a maximal mean discrepancy (MMD) loss \cite{gretton2012kernel, meng2023state} to move the conditional membership distribution far away from the uniform distribution $\mathcal{U}_\mathcal{N}$, which is depicted in Eq. \ref{eq:ln}.
\begin{equation}
\label{eq:ln}
\mathcal{L}_\mathcal{N}(q) = -\mathbb{E}_z\sum_{i=1}^n(q_i(z)-\frac{1}{n})^2
\end{equation}

\subsection{Maximum manifold capacity}
For $n$ manifolds embedded in a $d$ dimensional feature space, manifold capacity theory focuses on finding the maximal values of $\frac{n}{d}$ for which a random dichotomy can be separated by a hyperplane \cite{cover1965geometrical, gardner1988space}. MMCR proposed an SSL objective based on manifold capacity \cite{yerxa2024learning}, which can be written as the negative nuclear norm of the centroid matrix $C$, shown in Eq. \ref{eq:mmcr}.

\begin{equation}
\label{eq:mmcr}
\mathcal{L}_{MMCR} = -||C||_*
\end{equation}

$-||C||_*$ is minimized when the centroid vectors, which are the averages of normalized within-manifold vectors, maintain orthogonality to each other and reach the upper bound constrained to a maximum value of 1. The goal of learning with this objective is to develop effective and meaningful manifold representations, and a connection between this objective and maximizing the lower bound on mutual information was made by \cite{isik2023information}.

\subsection{Capacity as a regularizer}
In our proposed method, we utilize the negative nuclear norm as a regularization technique rather than an explicit learning goal. This approach is symbolized by C$^+$. Additionally, we introduce a hyper-parameter, $\epsilon$, to control the magnitude of the negative nuclear norm.

To rationalize our approach, we begin by examining the disparities between the transformations of input data and the multiple outputs generated by distinct network heads. A common issue identified in bootstrapped methods is the convergence of outputs from multiple heads towards similar values after extensive training \cite{meng2022improving}. This convergence limits the information gained from directly employing MVSSL on these outputs, and as a result, using multiple output heads leads to worse performance than using input transformations.

We claim that by integrating the negative nuclear norm with existing MVSSL methods, we can enhance the performance of multi-headed outputs without the need to introduce additional transformations. We also need to demonstrate that the mutual information $I[\phi_1(z)\; ; \;\phi_2(z)]$ is maximized when it is coupled with the InfoNCE loss, as the connections between the InfoNCE loss and other learning objectives are well-established.

\paragraph{Theorem 1.} \textit{Let $\phi_1$ and $\phi_2$ be two charts that map the latent embeddings $z$ to two different outputs. The true distribution is unknown and $p$ is a variational distribution. Then, we show that InfoNCE loss \cite{gutmann2010noise} maximizes the mutual information $I[\phi_1(z)\; ; \;\phi_2(z)]$.}

\begin{proof}
To show that, we start with Theorem 2.5 in \cite{isik2023information} stating that the model is able to achieve an lower bound on the negative nuclear norm with high probability based on perfect reconstruction and perfect uniformity. The mutual information must be larger or equal than the sum of the reconstruction and the entropy \cite{wang2020understanding, galvez2023role}:

\begin{equation}
\label{eq:mmcri}
    I[\;\phi_1(z)\; ; \;\phi_2(z)\;] \; \geq \; \mathbb{E}[\;\text{log} \; p(\phi_1(z) \; | \; \phi_2(z))\;] + H[\;\phi_1(z)\;]
\end{equation}

Here, the first term is the reconstruction and the second term is the entropy. The InfoNCE loss minimizes the negative logarithmic probability:

\begin{equation}
\label{eq:ncei}
    \mathcal{L}_{InfoNCE} = -\mathbb{E}[\;\text{log}\;\frac{d(\phi_1(z), \phi_2(z))}{\sum_{z'\in Z} d(\phi_1(z'),  \phi_2(z))}\;]
\end{equation}

Since $d(\phi_1(z), \phi_2(z))$ in the InfoNCE loss is an estimator of the density ratio $\frac{p(\phi_1(z)\;|\;\phi_2(z))}{p(\phi_1(z))}$, the InfoNCE loss maximizes $I[\phi_1(z)\; ; \;\phi_2(z)]$ following the lower bound in Eq. \ref{eq:mmcri}.
\end{proof}

\subsection{DeepInfoMax with capacity}
DeepInfoMax aims to achieve two essential objectives: a local-local objective ($\mathcal{L}_{Local}$) that depicts the interactions within local features, and a global-local objective ($\mathcal{L}_{Global}$) that represents the exchange of information between global structure and local details. These objectives are formalized in Eq. \ref{eq:gl} and \ref{eq:ll}, respectively, as detailed in \cite{hjelm2018learning, anand2019unsupervised, meng2023state}.

\begin{equation}
\mathcal{L}_{Global}=\sum^{H}\sum^{W}-\text{log}\frac{\text{exp}(bs1(x_t,x_{t+1}))}{\sum_{x_{t*} \in X}\text{exp}(bs1(x_t,x_{t*}))}
\label{eq:gl}
\end{equation}

\begin{equation}
\mathcal{L}_{Local}=\sum^{H}\sum^{W}-\text{log}\frac{\text{exp}(bs2(x_t,x_{t+1}))}{\sum_{x_{t*} \in X}\text{exp}(bs2(x_t,x_{t*}))}
\label{eq:ll}
\end{equation}

Given a set of sampled observations denoted by $X$, $x_t$ and $x_{t+1}$ are temporally adjacent pairs and $x_{t*}$ is a randomly selected sample within $X$. $H$ and $W$ are the height and width of the local feature map. The similarity between feature pairs relevant to $\mathcal{L}_{Global}$ and $\mathcal{L}_{Local}$ is calculated using the bilinear scoring functions $bs1$ and $bs2$, respectively.

In our approach, DIM-C$^+$, we aim to demonstrate that employing just the capacity regularizer is sufficient for acquiring satisfactory performance. To this end, we equip each chart with a dedicated two-layer multilayer perceptron (MLP) head. The global-local objective, $\mathcal{L}_{{Global}}$, is computed separately for each output by processing through their corresponding chart function $\phi_i$. During inference, we discard the entire MLP head, in contrast to the strategy employed by DIM-UA which selects the head with the highest membership probability.

Furthermore, to comprehensively evaluate the integration of UA and C$^+$, we also incorporate the capacity constraint with DIM-UA and name this as DIM-UAC$^+$. The loss for DIM-UAC$^+$ is shown in Eq. \ref{eq:uac}. Unlike DIM-UAC$^+$, the total loss for DIM-C$^+$ omits $\mathcal{L}_{\mathcal{N}}$ due to the absence of the membership probability, as described by Eq. \ref{eq:c}. The algorithmic distinction between DIM-C$^+$ and DIM-UAC$^+$ can be revealed upon examining Algorithm \ref{algo:dim-c} and Algorithm \ref{algo:dim-uac} presented in Appendix \ref{app:de}.

\begin{equation}
\mathcal{L}_{UAC^+} = \mathcal{L}_{Global} + \mathcal{L}_{Local} + \mathcal{L}_{\mathcal{N}} + \mathcal{L}_{MMCR}
\label{eq:uac}
\end{equation}

\begin{equation}
\mathcal{L}_{C^+} = \mathcal{L}_{Global} + \mathcal{L}_{Local} + \mathcal{L}_{MMCR}
\label{eq:c}
\end{equation}

In practice, we apply the $\mathcal{L}_{MMCR}$ objective selectively to only one of the model's outputs to prevent model collapse and the convergence issues previously mentioned. This technique is comparable to the use of the stop-gradient operation in SimSiam \cite{chen2021exploring}.

\section{Experimental Details}
The experiments use a single NVIDIA Tesla V100 GPU and an 8-core CPU, utilizing the PyTorch framework \cite{NEURIPS2019_9015}. The performance of DIM-C$^+$ is assessed alongside other SRL methods using the AtariARI benchmark across 19 games. For each game, we run independent experiments with five different seeds to obtain the averaged results. AtariARI categorizes state variables into five groups \cite{anand2019unsupervised}: agent localization, small object localization, other localization, miscellaneous, and score, clock, lives, and display information. By categorizing them, we are able to weigh the results of each component, as they all collectively contribute to the overall performance of the RL agents. To gather the necessary data for pretraining and subsequent probing, we deploy RL agents that execute a predefined number of steps under a random policy. Following the previous SRL workflow, we report linear probe accuracy and F1 scores for the downstream tasks.

The encoder undergoes pretraining before it is tasked with predicting an input image's ground truth via an auxiliary linear classifier. It is noteworthy that for linear probing in DIM-UAC$^+$, we retain the output head with the maximal membership probability, while for DIM-C$^+$, only the model backbone is kept during this phase. The MLP within DIM-C$^+$ uses a hidden layer of 2048 units, and we need to go through each head to compute the global objective, as delineated in Algorithm \ref{algo:dim-c} in Appendix \ref{app:de}, incurring additional computational costs during pretraining. Nevertheless, DIM-C$^+$ has greater efficiency at inference time when compared to DIM-UAC$^+$. The state-of-the-art method DIM-UA deploys four output heads, each with 4096 units. For a legitimate comparison based on performance, we use the same amount of output units and maintain identical hyper-parameters as much as possible across DIM-UA, DIM-UAC$^+$, and DIM-C$^+$.

For SimCLR and BT, we integrate the capacity regularizer analogously to the method employed by DIM-C$^+$. However, it is necessary to pair the outputs from each head across both views to compute the SimCLR and BT objectives pairwise. We evaluate their performances using the CIFAR10 and STL10 datasets for image classification tasks, adhering to the experimental protocols from \cite{tsai2021note}. Both SimCLR and BT, along with their C$^+$ variations that use eight output charts with 256 hidden units in each, maintain a consistent sum of 2048 output units in the MLP. The code snippet is shown in Algorithm \ref{algo:c} and details concerning the hyper-parameter selections are listed in Table \ref{tbl:parac} within Appendix \ref{app:de}.

\section{Results}
\subsection{Linear F1 scores}

\begin{table*}[t]
\caption{Linear F1 scores for each game averaged across categories}
\centering
\begin{tabular}{ c c c c c c c} 
 \hline
 Game &VAE &CPC &ST-DIM &DIM-UA &DIM-UAC$^+$&DIM-C$^+$\\\hline
Asteroids &0.36 &0.42& 0.49 & \textbf{0.5} & 0.49 $\pm$ 0.026& \textbf{0.5} $\pm$ 0.025\\
Bowling& 0.50 &0.90 &0.96 & 0.96& \textbf{0.97} $\pm$ 0.005 & 0.96 $\pm$ 0.011\\
Boxing & 0.20& 0.29& 0.58& 0.64 & 0.65 $\pm$ 0.022 &\textbf{0.69} $\pm$ 0.022\\
Breakout &0.57 &0.74 &0.88 & \textbf{0.9}& \textbf{0.9} $\pm$ 0.012 & 0.89 $\pm$ 0.014\\
Demon Attack & 0.26& 0.57 &0.69&0.74& 0.72 $\pm$ 0.024 &\textbf{0.75} $\pm$ 0.021\\
Freeway &0.01 & 0.47& 0.81 &0.86& 0.86 $\pm$ 0.014 & \textbf{0.95} $\pm$ 0.011\\
Frostbite &0.51 &0.76 &0.75 &0.75& 0.74 $\pm$ 0.01 & \textbf{0.78} $\pm$ 0.009\\
Hero &0.69&0.90 &0.93 &0.94& 0.94 $\pm$ 0.012 & \textbf{0.96} $\pm$ 0.011\\
Montezuma Revenge &0.38 & 0.75 &0.78&\textbf{0.84} &0.82 $\pm$ 0.01 & \textbf{0.84} $\pm$ 0.008\\
Ms Pacman &0.56 &0.65 &0.72 &0.76 & 0.75 $\pm$ 0.017&\textbf{0.77} $\pm$ 0.013\\
Pitfall &0.35 & 0.46& 0.60 &\textbf{0.73} & 0.71 $\pm$ 0.028&\textbf{0.73} $\pm$ 0.026\\
Pong & 0.09 &0.71 &0.81 & \textbf{0.85} &\textbf{0.85} $\pm$ 0.005 &0.84 $\pm$ 0.006\\
Private Eye &0.71 & 0.81& 0.91 &0.93 &0.92 $\pm$ 0.011 & \textbf{0.94} $\pm$ 0.01\\
Qbert & 0.49& 0.65 &0.73&0.79 & 0.79 $\pm$ 0.024 &\textbf{0.8} $\pm$ 0.022\\
Seaquest  &0.56 &0.66 &0.67 &0.69 &0.68 $\pm$ 0.01 & \textbf{0.71} $\pm$ 0.01\\
Space Invaders &0.52& 0.54& 0.57&0.62&0.59 $\pm$ 0.02 & \textbf{0.68} $\pm$ 0.013\\
Tennis &0.29 &0.60 &0.60 & 0.64 &0.64 $\pm$ 0.007 &\textbf{0.71} $\pm$ 0.009\\
Venture &0.38 &0.51 &0.58&0.58 &0.59 $\pm$ 0.02 &\textbf{0.61} $\pm$ 0.024\\
Video Pinball &0.45 &0.58 &0.61&0.62 & \textbf{0.67} $\pm$ 0.023 &0.66 $\pm$ 0.021\\\hline
Mean &0.41&0.63&0.72 & 0.75 & 0.75 $\pm$ 0.016 &\textbf{0.78} $\pm$ 0.015\\
\hline
\end{tabular}
\label{tbl:resf1o}
\end{table*}

In this section, we present the F1 scores of linear probe tasks for our experiments and compare DIM-C$^+$ with other SRL methods to validate the effectiveness of implementing capacity as a regularizer. The linear accuracy scores are shown in Table \ref{tbl:reso} within Appendix \ref{app:res}, as the values of accuracy scores and F1 scores are similar in our experiments.

Furthermore, we include the results of DIM-UAC$^+$ in the table, as we aim to examine the performance implications of integrating capacity with the UA theorem. The outcomes, delineated in Table \ref{tbl:resf1o}, are based on averages across each variable category. By aggregating state variables into distinct categories and averaging them, we adhere to the established SRL evaluation protocol and acknowledge the significance of each variable group. For the baseline results, we reference the results from \cite{anand2019unsupervised} for VAE, CPC, and ST-DIM, which all use 256 output units. Meanwhile, the results of DIM-UA are taken from \cite{meng2023state}, which use a total of 16384 output units and four heads. For a legitimate comparison to DIM-UA, both DIM-UAC$^+$ and DIM-C$^+$ use a total of 16384 output units, and allocate them into four heads, with each head comprising 4096 units.

In Table \ref{tbl:resf1o}, DIM-C$^+$ matches or exceeds the F1 scores of all other methods in 15 out of the 19 evaluated games, and it achieves higher F1 scores than the baseline DIM-UA in 13 of the 19 games. The mean F1 score of DIM-C$^+$ is 78\%, which is also the highest among all evaluated methods, overtaking the 75\% F1 score of the previous state-of-the-art method DIM-UA and indicating a significant improvement enabled by the capacity regularizer. In contrast, DIM-UAC$^+$ achieves equal or higher F1 scores compared to DIM-UA in 10 out of 19 games, yet only surpasses DIM-UA in four games. Both DIM-UA and DIM-UAC$^+$ share an average F1 score of 0.75, suggesting that the combination of a capacity regularizer with the UA theorem only yields marginal gains.

Notably, DIM-C$^+$ obtains significantly higher F1 scores than all other methods in Freeway, Space Invaders, and Tennis by margins of 9\%, 6\%, and 7\% respectively. These results indicate the strength of DIM-C$^+$ in games where other methods struggle to learn more useful representations. It was previously found out that other approaches can generate adequate representations for some seeds in Freeway, but fail for some other seeds, as they are prone to model collapse \cite{meng2023state}. Instances where DIM-C$^+$ underperforms typically occur in scenarios where prior methods have already established competent representations. This supports our hypothesis that utilizing capacity as a regularizer alone is sufficient to enhance model stability and learn more powerful and robust representations.

\subsection{Accuracy for image classification tasks}

\begin{table}[t]
\caption{Linear evaluation accuracy for SimCLRC$^+$ and BTC$^+$, with 2048 output units used for the baseline. For  C$^+$ methods, eight heads with 256 output units for each head are used.}
\begin{center}
\begin{tabular}{c c c c}
\hline
\textbf{Method}&\multicolumn{2}{c}{\textbf{Dataset}}\\
\cline{2-3}
\textbf{} & CIFAR10& STL10\\
\hline
SimCLR & 0.911 & 0.889 \\

C$^+$  & \textbf{0.913} & \textbf{0.912} \\
\hline
BT & 0.905 & 0.88 \\

C$^+$  & \textbf{0.922} & \textbf{0.906} \\
\hline
\end{tabular}
\label{tbl:ic}
\end{center}
\end{table}

In Table \ref{tbl:ic}, SimCLR achieves an accuracy of 91.1\% on CIFAR10. Meanwhile, SimCLRC$^+$ surpasses this slightly by having an improved accuracy of 91.3\%, also exceeding the baseline accuracy 88.6\% of SimCLR under the UA paradigm \cite{meng2023state}. The enhancement is more pronounced in STL10, where SimCLR$^+$ records an accuracy of 91.2\%, substantially higher than the 88.9\% achieved by SimCLR. Meanwhile, BTC$^+$ consistently outperforms BT by a large margin on both the CIFAR10 and STL10 datasets, showing accuracies of 92.2\% vs. 90.5\% and 90.6\% vs. 88\%, respectively.

The experiments on image classification tasks demonstrate that our proposed paradigm exhibits versatility, enhancing performances on SimCLR and BT, and is not confined solely to DIM-C$^+$. Secondly, its effectiveness is not affected by the size of the model, since ResNet50 is used as the backbone here, much larger than the smaller model with three convolutional layers in DIM-C$^+$. Therefore, capacity as a regularizer can yield significant advancements and it may serve as a universal strategy better than UA to effectively forge manifold representations within the SSL domain.

\subsection{Sensitivity to hyper-parameters}
\begin{figure}[t]
    \centering
    \includegraphics[width=0.8\linewidth]{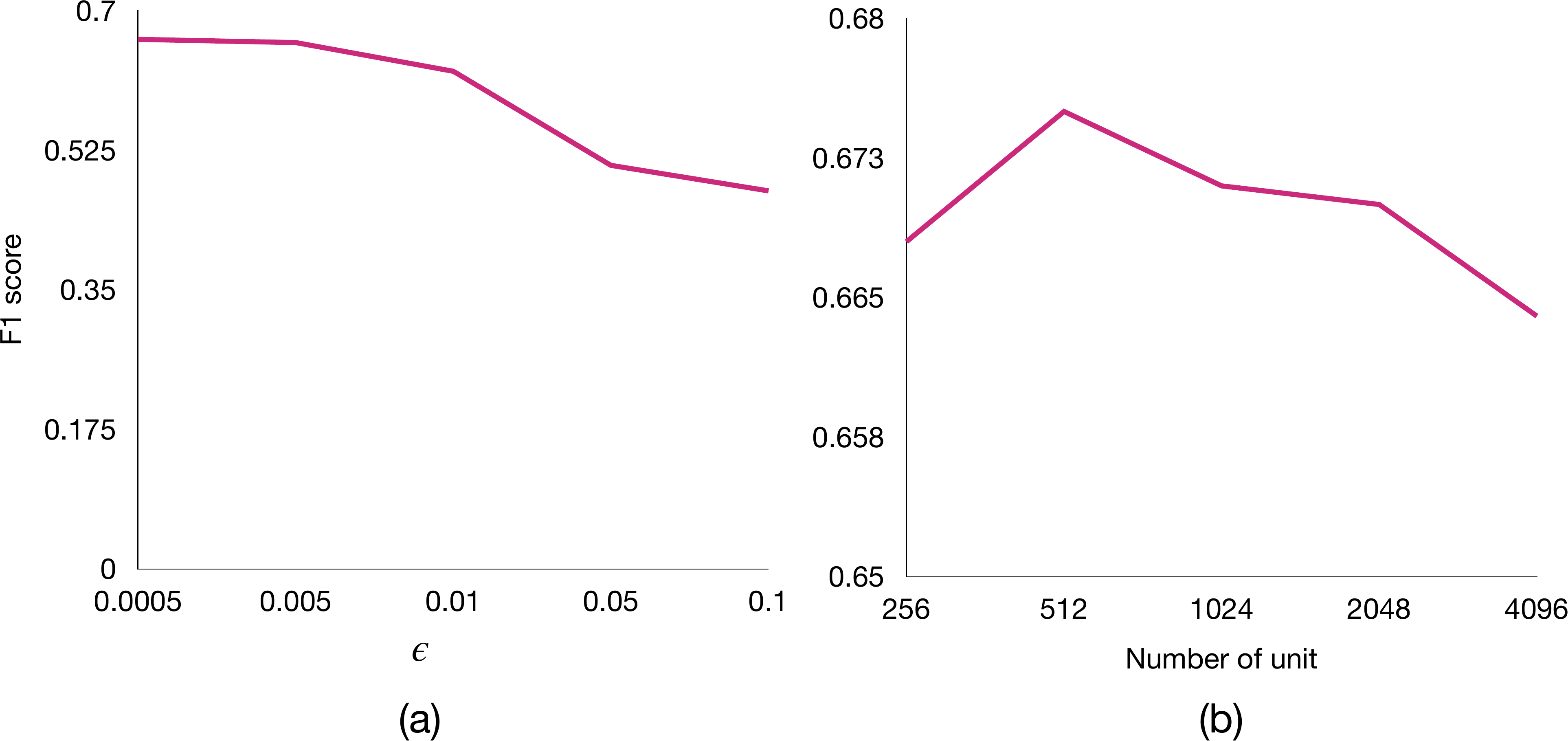}
    \caption{The F1 scores for different values of $\epsilon$ in (a) and different numbers of output units in (b). All other hyper-parameters are kept the same. Each value is an average of the F1 scores from five independent experiments.}
\label{fig:units}
\end{figure}

Figure \ref{fig:units}(a) shows the results of the F1 scores with different $\epsilon$ values on the game Video Pinball, utilizing the same hyper-parameters listed in Table \ref{tbl:resf1o}. We observe a clear trend, where the F1 score diminishes as $\epsilon$ increases. For the base value of $\epsilon$ set at 0.0005, we attain a result of 0.664, and this score progressively declines, hitting the lowest at 0.474 when $\epsilon$ is increased to 0.1. This downturn suggests that larger $\epsilon$ values cause negative effects, which is anticipated, considering that a large $\epsilon$ results in a disproportionate share of the composite loss being attributed to the MMCR loss, diminishing the relative influence of both the global-local and local-local losses within the pretraining.

Nonetheless, the disparity between an $\epsilon$ of 0.005 and 0.0005 is relatively minimal. An F1 score of 0.66 is observed when $\epsilon$ is set to 0.005, implying that performance remains relatively stable when $\epsilon$ is maintained at a reasonably low level, while adopting an excessively high $\epsilon$ could compromise the performance.

Figure \ref{fig:units}(b) illustrates how the F1 scores for Video Pinball scale with the number of output units for DIM-C$^+$. At first glance, it may appear somewhat counter-intuitive that the highest F1 score achieved is around 0.675 with just 512 units, which surpasses the score of 0.66 reported in Table \ref{tbl:resf1o} with 4096 units. Contrary to the reported results in the previous work \cite{meng2023state}, a greater number of output units does not correlate with improved performance in DIM-C$^+$, and our reported scores for Video Pinball with 4096 units seem to sit at the lower end of the scores observed in this hyper-parameter sweep. A plausible reason for the optimal score being obtained with using $4\times512$ units is that the hidden layer in our MLP also has 2048 units, and the effect of scaling the units in hidden layers is left for future study. This finding is consistent with results presented in Table \ref{tbl:ic}, which suggest that using 256 units per head across eight heads is adequate for obtaining improved results in image classification tasks.

However, given that all F1 scores in Figure \ref{fig:units}(b) fluctuates marginally around 0.66, we conclude that our algorithm exhibits low sensitivity to variations in the number of output units. The discrepancy between this conclusion and the observations from DIM-UA is likely stemmed from the network head structure within our method. Our enhanced expressiveness can be attributed to adopting an MLP in DIM-C$^+$, as opposed to a single linear layer following ReLU activation in DIM-UA. In the case of DIM-UA, a substantial count of output units is needed due to the limited expressiveness when only a single layer with a small number of units is utilized. This contrasts with the approach within our paradigm, where the necessity for an extensive number of output units is mitigated by using hidden layers, suggesting that MLPs may be pivotal for producing high-quality representations. As a result, the overall performance should not exhibit substantial variance in response to the number of output units in an MLP.

\begin{figure}[t]
    \centering
    \includegraphics[width=0.4\linewidth]{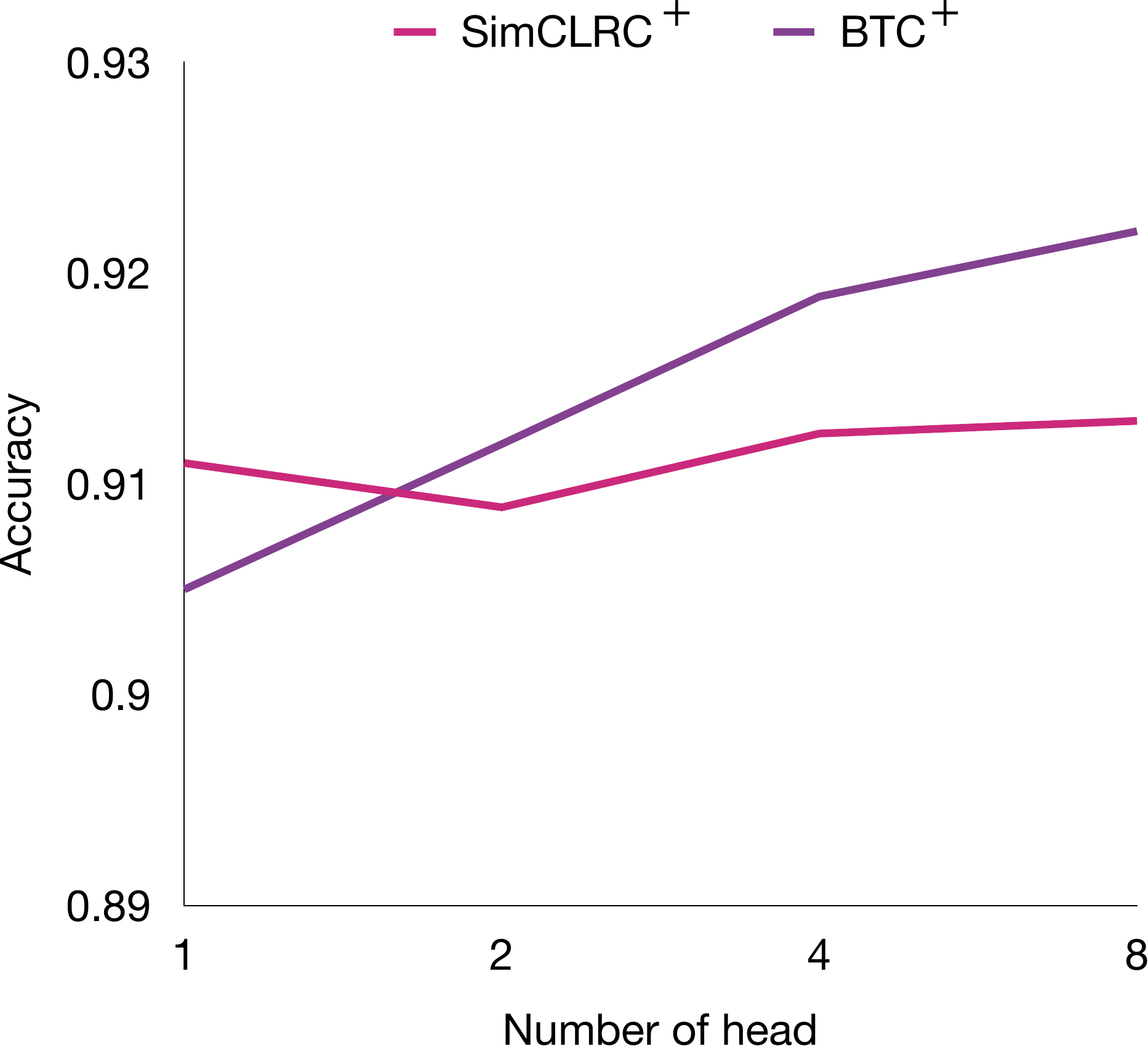}
    \caption{The accuracy scores on CIFAR10 for choosing different numbers of heads while keeping the total number of output units fixed (i.e., $1\times2048$, $2\times1024$, $4\times512$, $8\times256$).}
\label{fig:bacheads}
\end{figure}

Figure \ref{fig:bacheads} presents the impact of varying the number of heads while fixing the total number of units for image classification tasks on the CIFAR10 dataset. Here, SimCLRC$^+$ and BTC$^+$ with a single head are equivalent to SimCLR and BT. The results highlight that both SimCLRC$^+$ and BTC$^+$ exhibit improved performance when the number of heads is sufficiently large, though the enhancement is notably more pronounced and consistent for BTC$^+$. On the other hand, SimCLR$^+$ experiences a slight decline when using two output heads, and the overall difference in performance is not as significant. Nevertheless, it is premature to conclude that SimCLR does not benefit from a capacity regularizer, as there are substantial gains for SimCLRC$^+$ in Table \ref{tbl:ic} on the STL10 dataset.

\section{Discussion}
We have shown that utilizing capacity as a regularizer can enhance manifold representations, leading to improved performance in both SRL and MVSSL algorithms. Nonetheless, applying C$^+$ directly to the UA paradigm does not yield significantly improved results. This disparity may stem from the utilization of membership probability in UA, which could have potentially restricted the diversity of chart embeddings without carefully designed parameter annealing schedules. Meanwhile, the theoretical framework of UA relies on the convexity presumption. This assumption hinders UA from leveraging the MLP's capabilities and the universal approximation theorem, as it was previously shown that using more hidden layers in output heads had not further improved the results of DIM-UA \cite{meng2023state}. Our image classification experiments have also surpassed prior methods that rely on the encoding space with a membership probability \cite{korman2021self}.

The considerable improvement in mean F1 scores as depicted in Table \ref{tbl:resf1o} underscores the potential of DIM-C$^+$ to significantly influence future research in learning state representations for RL. Similarly, the introduction of capacity as a regularizer to other existing MVSSL methods also has the potential to substantially enhance their performances. Future research may exploit  DIM-C$^+$ to learn useful representations in pretraining for a variety of games or control tasks, and may develop more sophisticated representations that also aim to reduce computational costs.

\section{Limitations}
Our work aims to mitigate the considerable computational demands associated with optimizing the MMCR objective across multiple views. By combining MMCR with other MVSSL objectives, the necessity for more than two views is eliminated. Nonetheless, it should be noted that the computational load during pretraining for our method remains higher compared to employing solely MVSSL objectives.

Both the C$^+$ and UA paradigms have the capacity to represent a manifold using more encoding dimensions while increasing the number of output heads. However, our experimental insights regarding the impact of expanding heads and dimensions are constrained, as extensive scaling could drastically inflate the computational overhead.

While the C$^+$ paradigm is adaptable to a wide range of MVSSL methods, the selection of the hyper-parameter $\epsilon$ to moderate the MMCR loss is somewhat nuanced. A universal $\epsilon$ choice across all MVSSL methods may not be practical. One potentially effective heuristic is to monitor the values for each MVSSL loss and adjust the associated MMCR loss proportionately.

\section*{Acknowledgments}
This work was performed on the resources from the Centre for Artificial Intelligence Research, University of Agder, and the Department for Research Computing at USIT, University of Oslo.

\bibliographystyle{abbrv}
{\small{
\bibliography{main}}}

\begin{thebibliography}{10}

\bibitem{alberti2023manifold}
G.~S. Alberti, J.~Hertrich, M.~Santacesaria, and S.~Sciutto.
\newblock Manifold learning by mixture models of vaes for inverse problems.
\newblock {\em arXiv preprint arXiv:2303.15244}, 2023.

\bibitem{anand2019unsupervised}
A.~Anand, E.~Racah, S.~Ozair, Y.~Bengio, M.-A. C{\^o}t{\'e}, and R.~D. Hjelm.
\newblock Unsupervised state representation learning in atari.
\newblock {\em Advances in neural information processing systems}, 32, 2019.

\bibitem{bachman2019learning}
P.~Bachman, R.~D. Hjelm, and W.~Buchwalter.
\newblock Learning representations by maximizing mutual information across views.
\newblock {\em Advances in neural information processing systems}, 32, 2019.

\bibitem{bojanowski2017unsupervised}
P.~Bojanowski and A.~Joulin.
\newblock Unsupervised learning by predicting noise.
\newblock In {\em International Conference on Machine Learning}, pages 517--526. PMLR, 2017.

\bibitem{caron2018deep}
M.~Caron, P.~Bojanowski, A.~Joulin, and M.~Douze.
\newblock Deep clustering for unsupervised learning of visual features.
\newblock In {\em Proceedings of the European conference on computer vision (ECCV)}, pages 132--149, 2018.

\bibitem{caron2019unsupervised}
M.~Caron, P.~Bojanowski, J.~Mairal, and A.~Joulin.
\newblock Unsupervised pre-training of image features on non-curated data.
\newblock In {\em Proceedings of the IEEE/CVF International Conference on Computer Vision}, pages 2959--2968, 2019.

\bibitem{caron2020unsupervised}
M.~Caron, I.~Misra, J.~Mairal, P.~Goyal, P.~Bojanowski, and A.~Joulin.
\newblock Unsupervised learning of visual features by contrasting cluster assignments.
\newblock {\em Advances in neural information processing systems}, 33:9912--9924, 2020.

\bibitem{chen2020simple}
T.~Chen, S.~Kornblith, M.~Norouzi, and G.~Hinton.
\newblock A simple framework for contrastive learning of visual representations.
\newblock In {\em International conference on machine learning}, pages 1597--1607. PMLR, 2020.

\bibitem{chen2020improved}
X.~Chen, H.~Fan, R.~Girshick, and K.~He.
\newblock Improved baselines with momentum contrastive learning.
\newblock {\em arXiv preprint arXiv:2003.04297}, 2020.

\bibitem{chen2021exploring}
X.~Chen and K.~He.
\newblock Exploring simple siamese representation learning.
\newblock In {\em Proceedings of the IEEE/CVF conference on computer vision and pattern recognition}, pages 15750--15758, 2021.

\bibitem{cover1965geometrical}
T.~M. Cover.
\newblock Geometrical and statistical properties of systems of linear inequalities with applications in pattern recognition.
\newblock {\em IEEE transactions on electronic computers}, pages 326--334, 1965.

\bibitem{eysenbach2022contrastive}
B.~Eysenbach, T.~Zhang, S.~Levine, and R.~R. Salakhutdinov.
\newblock Contrastive learning as goal-conditioned reinforcement learning.
\newblock {\em Advances in Neural Information Processing Systems}, 35:35603--35620, 2022.

\bibitem{galvez2023role}
B.~R. G{\'a}lvez, A.~Blaas, P.~Rodr{\'\i}guez, A.~Golinski, X.~Suau, J.~Ramapuram, D.~Busbridge, and L.~Zappella.
\newblock The role of entropy and reconstruction in multi-view self-supervised learning.
\newblock In {\em International Conference on Machine Learning}, pages 29143--29160. PMLR, 2023.

\bibitem{gardner1988space}
E.~Gardner.
\newblock The space of interactions in neural network models.
\newblock {\em Journal of physics A: Mathematical and general}, 21(1):257, 1988.

\bibitem{gretton2012kernel}
A.~Gretton, K.~M. Borgwardt, M.~J. Rasch, B.~Sch{\"o}lkopf, and A.~Smola.
\newblock A kernel two-sample test.
\newblock {\em The Journal of Machine Learning Research}, 13(1):723--773, 2012.

\bibitem{grill2020bootstrap}
J.-B. Grill, F.~Strub, F.~Altch{\'e}, C.~Tallec, P.~Richemond, E.~Buchatskaya, C.~Doersch, B.~Avila~Pires, Z.~Guo, M.~Gheshlaghi~Azar, et~al.
\newblock Bootstrap your own latent-a new approach to self-supervised learning.
\newblock {\em Advances in neural information processing systems}, 33:21271--21284, 2020.

\bibitem{gutmann2010noise}
M.~Gutmann and A.~Hyv{\"a}rinen.
\newblock Noise-contrastive estimation: A new estimation principle for unnormalized statistical models.
\newblock In {\em Proceedings of the thirteenth international conference on artificial intelligence and statistics}, pages 297--304. JMLR Workshop and Conference Proceedings, 2010.

\bibitem{he2020momentum}
K.~He, H.~Fan, Y.~Wu, S.~Xie, and R.~Girshick.
\newblock Momentum contrast for unsupervised visual representation learning.
\newblock In {\em Proceedings of the IEEE/CVF conference on computer vision and pattern recognition}, pages 9729--9738, 2020.

\bibitem{hjelm2018learning}
R.~D. Hjelm, A.~Fedorov, S.~Lavoie-Marchildon, K.~Grewal, P.~Bachman, A.~Trischler, and Y.~Bengio.
\newblock Learning deep representations by mutual information estimation and maximization.
\newblock {\em arXiv preprint arXiv:1808.06670}, 2018.

\bibitem{ioffe2015batch}
S.~Ioffe and C.~Szegedy.
\newblock Batch normalization: Accelerating deep network training by reducing internal covariate shift.
\newblock In {\em International conference on machine learning}, pages 448--456. pmlr, 2015.

\bibitem{isik2023information}
B.~Isik, V.~Lecomte, R.~Schaeffer, Y.~LeCun, M.~Khona, R.~Shwartz-Ziv, S.~Koyejo, and A.~Gromov.
\newblock An information-theoretic understanding of maximum manifold capacity representations.
\newblock In {\em UniReps: the First Workshop on Unifying Representations in Neural Models}, 2023.

\bibitem{jonschkowski2015learning}
R.~Jonschkowski and O.~Brock.
\newblock Learning state representations with robotic priors.
\newblock {\em Autonomous Robots}, 39(3):407--428, 2015.

\bibitem{kadeethum2022reduced}
T.~Kadeethum, F.~Ballarin, D.~O’malley, Y.~Choi, N.~Bouklas, and H.~Yoon.
\newblock Reduced order modeling for flow and transport problems with barlow twins self-supervised learning.
\newblock {\em Scientific Reports}, 12(1):20654, 2022.

\bibitem{kingma2013auto}
D.~P. Kingma and M.~Welling.
\newblock Auto-encoding variational bayes.
\newblock {\em arXiv preprint arXiv:1312.6114}, 2013.

\bibitem{korman2018autoencoding}
E.~O. Korman.
\newblock Autoencoding topology.
\newblock {\em arXiv preprint arXiv:1803.00156}, 2018.

\bibitem{korman2021atlas}
E.~O. Korman.
\newblock Atlas based representation and metric learning on manifolds.
\newblock {\em arXiv preprint arXiv:2106.07062}, 2021.

\bibitem{korman2021self}
E.~O. Korman.
\newblock Self-supervised representation learning on manifolds.
\newblock In {\em ICLR 2021 Workshop on Geometrical and Topological Representation Learning}, 2021.

\bibitem{laskin2020curl}
M.~Laskin, A.~Srinivas, and P.~Abbeel.
\newblock Curl: Contrastive unsupervised representations for reinforcement learning.
\newblock In {\em International Conference on Machine Learning}, pages 5639--5650. PMLR, 2020.

\bibitem{lesort2018state}
T.~Lesort, N.~D{\'\i}az-Rodr{\'\i}guez, J.-F. Goudou, and D.~Filliat.
\newblock State representation learning for control: An overview.
\newblock {\em Neural Networks}, 108:379--392, 2018.

\bibitem{macqueen1967some}
J.~MacQueen et~al.
\newblock Some methods for classification and analysis of multivariate observations.
\newblock In {\em Proceedings of the fifth Berkeley symposium on mathematical statistics and probability}, volume~1, pages 281--297. Oakland, CA, USA, 1967.

\bibitem{makhzani2015adversarial}
A.~Makhzani, J.~Shlens, N.~Jaitly, I.~Goodfellow, and B.~Frey.
\newblock Adversarial autoencoders.
\newblock {\em arXiv preprint arXiv:1511.05644}, 2015.

\bibitem{meng2022improving}
L.~Meng, M.~Goodwin, A.~Yazidi, and P.~Engelstad.
\newblock Improving the diversity of bootstrapped dqn by replacing priors with noise.
\newblock {\em IEEE Transactions on Games}, 2022.

\bibitem{meng2023unsupervised}
L.~Meng, M.~Goodwin, A.~Yazidi, and P.~Engelstad.
\newblock Unsupervised state representation learning in partially observable atari games.
\newblock In {\em International Conference on Computer Analysis of Images and Patterns}, pages 212--222. Springer, 2023.

\bibitem{meng2023state}
L.~Meng, M.~Goodwin, A.~Yazidi, and P.~E. Engelstad.
\newblock State representation learning using an unbalanced atlas.
\newblock In {\em The Twelfth International Conference on Learning Representations}, 2023.

\bibitem{oord2018representation}
A.~v.~d. Oord, Y.~Li, and O.~Vinyals.
\newblock Representation learning with contrastive predictive coding.
\newblock {\em arXiv preprint arXiv:1807.03748}, 2018.

\bibitem{NEURIPS2019_9015}
A.~Paszke, S.~Gross, F.~Massa, A.~Lerer, J.~Bradbury, G.~Chanan, T.~Killeen, Z.~Lin, N.~Gimelshein, L.~Antiga, A.~Desmaison, A.~Kopf, E.~Yang, Z.~DeVito, M.~Raison, A.~Tejani, S.~Chilamkurthy, B.~Steiner, L.~Fang, J.~Bai, and S.~Chintala.
\newblock Pytorch: An imperative style, high-performance deep learning library.
\newblock In {\em Advances in Neural Information Processing Systems 32}, pages 8024--8035. Curran Associates, Inc., 2019.

\bibitem{pitelis2013learning}
N.~Pitelis, C.~Russell, and L.~Agapito.
\newblock Learning a manifold as an atlas.
\newblock In {\em Proceedings of the IEEE Conference on Computer Vision and Pattern Recognition}, pages 1642--1649, 2013.

\bibitem{stooke2021decoupling}
A.~Stooke, K.~Lee, P.~Abbeel, and M.~Laskin.
\newblock Decoupling representation learning from reinforcement learning.
\newblock In {\em International Conference on Machine Learning}, pages 9870--9879. PMLR, 2021.

\bibitem{tsai2021note}
Y.-H.~H. Tsai, S.~Bai, L.-P. Morency, and R.~Salakhutdinov.
\newblock A note on connecting barlow twins with negative-sample-free contrastive learning.
\newblock {\em arXiv preprint arXiv:2104.13712}, 2021.

\bibitem{wang2020understanding}
T.~Wang and P.~Isola.
\newblock Understanding contrastive representation learning through alignment and uniformity on the hypersphere.
\newblock In {\em International conference on machine learning}, pages 9929--9939. PMLR, 2020.

\bibitem{yerxa2024learning}
T.~Yerxa, Y.~Kuang, E.~Simoncelli, and S.~Chung.
\newblock Learning efficient coding of natural images with maximum manifold capacity representations.
\newblock {\em Advances in Neural Information Processing Systems}, 36, 2024.

\bibitem{zbontar2021barlow}
J.~Zbontar, L.~Jing, I.~Misra, Y.~LeCun, and S.~Deny.
\newblock Barlow twins: Self-supervised learning via redundancy reduction.
\newblock In {\em International Conference on Machine Learning}, pages 12310--12320. PMLR, 2021.

\bibitem{zhu2022tico}
J.~Zhu, R.~M. Moraes, S.~Karakulak, V.~Sobol, A.~Canziani, and Y.~LeCun.
\newblock Tico: Transformation invariance and covariance contrast for self-supervised visual representation learning.
\newblock {\em arXiv preprint arXiv:2206.10698}, 2022.

\end{thebibliography}

\appendix
\section{Additional Details}
\label{app:de}

Table \ref{tbl:para} illustrates the hyper-parameter settings for DIM-UA. Both DIM-UAC$^+$ and DIM-C$^+$ use those same parameters, except that there is no $\tau$ in DIM-C$^+$. We also choose the same value, 0.0005 of $\epsilon$ for both DIM-UAC$^+$ and DIM-C$^+$. The architecture of the neural network in DIM-UA is illustrated in Table \ref{nn:3}.

\begin{table}[ht]
\caption{Parameter settings for DIM-UA}
\centering
\begin{tabular}{c c} 
\hline
Parameter & Value \\\hline
Input size & 160 $\times$ 210 \\
Batch size & 64\\
Epochs & 100\\
$\tau$ & 0.1\\
Optimizer & Adam\\
Learning rate & 3e$^{-4}$\\
Pretraining steps & 8e$^4$ \\
Linear training steps & 3.5e$^4$\\
Linear testing steps & 1e$^4$\\
\hline
\end{tabular}
\label{tbl:para}
\end{table}

\begin{table}[ht]
\caption{Overview of the DIM-UA architecture}
\centering
\begin{tabular}{c} 
\hline
Input \\
$\downarrow$\\
Conv with kernel 8, stride 4,  32 channels\\
$\downarrow$\\
ReLU \\
$\downarrow$\\
Conv with kernel 4, stride 2,  64 channels\\
$\downarrow$\\
ReLU \\
$\downarrow$\\
Conv with kernel 4, stride 2,  128 channels\\
$\downarrow$\\
ReLU \\
$\downarrow$\\
Conv with kernel 3, stride 1,  64 channels\\
$\downarrow$\\
ReLU \\
$\downarrow$\\
Flattened\\
\hline
\end{tabular}
\label{nn:3}
\end{table}

\begin{table}[ht]
\caption{Default parameter settings for BT and SimCLR}
\centering
\begin{tabular}{c c c } 
\hline
Parameter & Pretrain & Linear \\\hline
Batch size & 128 & 512\\
Epochs & 1000 & 200\\
$\tau$ (SimCLR)  &\multicolumn{2}{c}{0.5} \\
$\lambda$ (BT)  &\multicolumn{2}{c}{0.005} \\
$\epsilon$  &\multicolumn{2}{c}{0.005} \\
Optimizer & \multicolumn{2}{c}{Adam}\\
Learning rate &\multicolumn{2}{c}{1e$^{-3}$} \\
Weight decay  &\multicolumn{2}{c}{1e$^{-6}$}\\
MLP &\multicolumn{2}{c}{2048 -- 2048}\\
\hline
\end{tabular}
\label{tbl:parac}
\end{table}

 \begin{algorithm}[ht]
\SetAlgoLined
    \PyComment{$\mathrm{out}$, $\mathrm{fm}$, $\mathrm{mp}$: network output, feature map, membership probability} \\
    \PyComment{$\mathrm{c1,\; c2}$: linear layers} \\
    \PyComment{B, D: batch size, hidden units} \\
    \PyComment{H, W, C: height, width, channels of feature map} \\
    \PyComment{$\mathrm{ua}$, $\mathrm{mmcr}$, $\mathrm{cross\_entropy}$: ua, mmcr, cross entropy losses}\\
    \PyComment{} \\
    \PyCode{for $x_t,\;x_{t+1}$ in batch:} \PyComment{$x_t$, $x_{t+1}$ in a batch} \\
    \Indp   % start indent
        \PyComment{network outputs} \\
        \PyCode{$o_t,\; y_t,\; y_{t+1},\; q_t, \; q_{t+1}\; =\; \mathrm{out}(x_t), \;\mathrm{fm}(x_{t}),\; \mathrm{fm}(x_{t+1}),\; \mathrm{mp}(x_t),\; \mathrm{mp}(x_{t+1})$}\\
        \PyCode{$loss \;=\; -0.05\;*\;(\mathrm{ua}(q_t)\;+\;\mathrm{ua}(q_{t+1}))$} \PyComment{ua loss}\\
        \PyCode{$loss \;+=\; 0.0005\;*\;\mathrm{mmcr}(\mathrm{norm}(o_t))$} \PyComment{mmcr loss}\\
        \PyCode{$loss_g\; = \;0,\; loss_l\; = \;0$} \PyComment{global-local and local-local losses}\\
        \PyComment{for each point in feature map} \\
        \PyCode{for $h$ in $\mathrm{range}(H)$}: \\
        \Indp   % start indent
        \PyCode{for $w$ in $\mathrm{range}(W)$}: \\
        \Indp   % start indent
            \PyComment{global-local loss} \\
            \PyCode{$logits_g\; = \; \mathrm{matmul}(\mathrm{c1}(\mathrm{mean}(o_t)),\; y_{t+1}[:,\;h,\;w,\; :].\mathrm{t}())$} \\
            \PyCode{$loss_g\; += \; \mathrm{cross\_entropy}(logits_g,\; \mathrm{range}(B))$} \\
            \PyComment{local-local loss} \\
            \PyCode{$logits_l\; = \; \mathrm{matmul}(\mathrm{c2}(y_{t}[:,\;h,\;w,\; :]),\; y_{t+1}[:,\;h,\;w,\; :].\mathrm{t}())$} \\
            \PyCode{$loss_l \; += \; \mathrm{cross\_entropy}(logits_l,\; \mathrm{range}(B))$} \\
        \Indm % end indent
        \Indm
        \PyCode{$loss_g\; /= \;H*W, \;loss_l\; /= \;H*W$} \PyComment{divided by feature map size}\\
        \PyCode{$loss \;+=\; loss_g \; + \; loss_l$}
        \PyComment{final loss} \\
    \Indm % end indent, must end with this, else all the below text will be indented
\caption{Pytorch-style code snippet for DIM-UAC$^+$.}
\label{algo:dim-uac}
\end{algorithm}

 \begin{algorithm}[ht]
\SetAlgoLined
    \PyComment{$\mathrm{out}$, $\mathrm{fm}$: network output, feature map} \\
    \PyComment{$\mathrm{c1,\; c2}$: linear layers} \\
    \PyComment{B, N, D: batch size, number of heads, hidden units} \\
    \PyComment{H, W, C: height, width, channels of feature map} \\
    \PyComment{$\mathrm{mmcr}$, $\mathrm{cross\_entropy}$: mmcr, cross entropy losses}\\
    \PyComment{} \\
    \PyCode{for $x_t,\;x_{t+1}$ in batch:} \PyComment{$x_t$, $x_{t+1}$ in a batch} \\
    \Indp   % start indent
        \PyComment{network outputs} \\
        \PyCode{$o_t,\; y_t,\; y_{t+1}\; =\; \mathrm{out}(x_t), \;\mathrm{fm}(x_{t}),\; \mathrm{fm}(x_{t+1})$}\\
        \PyCode{$loss \;=\; 0.0005\;*\;\mathrm{mmcr}(\mathrm{norm}(o_t))$} \PyComment{mmcr loss}\\
        \PyCode{$loss_g\; = \;0,\; loss_l\; = \;0$} \PyComment{global-local and local-local losses}\\
        \PyComment{for each point in feature map} \\
        \PyCode{for $h$ in $\mathrm{range}(H)$}: \\
        \Indp   % start indent
        \PyCode{for $w$ in $\mathrm{range}(W)$}: \\
        \Indp   % start indent
            \PyComment{global-local loss} \\
            \PyCode{for $n$ in $\mathrm{range}(N)$}: \\
            \Indp   % start indent
            \PyCode{$logits_g\; = \; \mathrm{matmul}(\mathrm{c1}(o_t[:,\;n,\; :]),\; y_{t+1}[:,\;h,\;w,\; :].\mathrm{t}())$} \\
            \PyCode{$loss_g\; += \; \mathrm{cross\_entropy}(logits_g,\; \mathrm{range}(B))\;/\;N$} \\
            \Indm
            \PyComment{local-local loss} \\
            \PyCode{$logits_l\; = \; \mathrm{matmul}(\mathrm{c2}(y_{t}[:,\;h,\;w,\; :]),\; y_{t+1}[:,\;h,\;w,\; :].\mathrm{t}())$} \\
            \PyCode{$loss_l \; += \; \mathrm{cross\_entropy}(logits_l,\; \mathrm{range}(B))$} \\
        \Indm % end indent
        \Indm
        \PyCode{$loss_g\; /= \;H*W, \;loss_l\; /= \;H*W$} \PyComment{divided by feature map size}\\
        \PyCode{$loss \;+=\; loss_g \; + \; loss_l$}
        \PyComment{final loss} \\
    \Indm % end indent, must end with this, else all the below text will be indented
\caption{Pytorch-style code snippet for DIM-C$^+$.}
\label{algo:dim-c}
\end{algorithm}

 \begin{algorithm}[ht]
\SetAlgoLined
    \PyComment{$\mathrm{out}$: network output} \\
    \PyComment{B, N: batch size, number of heads} \\
    \PyComment{$\mathrm{mmcr}$, $\mathrm{custom}$: mmcr loss, customized loss}\\
    \PyComment{} \\
    \PyCode{for $x,\;x'$ in batch:} \PyComment{$x'$, $x'$ in a batch} \\
    \Indp   % start indent
        \PyComment{network outputs} \\
        \PyCode{$o,\; o'\; =\; \mathrm{out}(x), \;\mathrm{out}(x')$}\\
        \PyCode{$loss \;=\; 0.005\;*\;\mathrm{mmcr}(o)$} \PyComment{mmcr loss}\\
        \PyComment{for each chart} \\
        \PyCode{for $i$ in $\mathrm{range}(N)$}: \\
        \Indp   % start indent
        \PyCode{for $j$ in $\mathrm{range}(N)$}: \\
        \Indp   % start indent
            \PyCode{$loss\; += \; \mathrm{custom}(o[:,\;i,\; :],\; o'[:,\;j,\; :])\;/\; N^2$} \\
        \Indm % end indent
        \Indm
    \Indm % end indent, must end with this, else all the below text will be indented
\caption{Pytorch-style code snippet for BT and SimCLR with C$^+$.}
\label{algo:c}
\end{algorithm}

Table \ref{tbl:parac} shows the parameter settings for BT and SimCLR, following the customs from \cite{chen2020simple, tsai2021note}. Nonetheless, we choose a fixed $\lambda$ of 0.005 since the benefit of scaling $\lambda$ according to the number of output units is not observed in our experiment. ResNet-50 is used as the backbone for this experiment. The output head consists of a two-layer MLP with 2048 hidden units. Between those two layers, batch normalization \cite{ioffe2015batch} and ReLU activation are inserted. 

The Pytorch-style code snippet for DIM-UAC$^+$ is shown in Algorithm \ref{algo:dim-uac}, whereas the Pytorch-style code snippet for DIM-C$^+$ is shown in Algorithm \ref{algo:dim-c}. The underlying algorithm for the BTC$^+$ and SimCLRC$^+$ methods is also depicted through the code snippet in Algorithm \ref{algo:c}.

\section{Supplemental Results}
\label{app:res}

\begin{table*}[ht]
\caption{Linear accuracy for each game averaged across categories}
\centering
\begin{tabular}{ c c c c c c c} 
 \hline
 
Game & VAE & CPC & ST-DIM &DIM-UA &DIM-UAC$^+$&DIM-C$^+$\\\hline
Asteroids &0.41 &0.48 &0.52&0.53 &0.52 $\pm$ 0.027 &\textbf{0.54} $\pm$ 0.025\\
Bowling & 0.56  & 0.90 &0.96 &0.96 & \textbf{0.97} $\pm$ 0.004 &0.96 $\pm$ 0.011\\
Boxing &0.23&  0.32 &0.59& 0.64 &0.65 $\pm$ 0.022 &\textbf{0.7} $\pm$ 0.021\\
Breakout & 0.61 &0.75& 0.89 &\textbf{0.91} &\textbf{0.91} $\pm$ 0.011 & 0.9 $\pm$ 0.012\\
Demon Attack &0.31  &0.58 &0.70 &0.74 &0.73 $\pm$ 0.02 &\textbf{0.75} $\pm$ 0.017\\
Freeway & 0.07& 0.49 &0.82&0.86&0.87 $\pm$ 0.012&\textbf{0.96} $\pm$ 0.008\\
Frostbite &0.54  &0.76 &0.75&0.75 &0.74 $\pm$ 0.01 &\textbf{0.79} $\pm$ 0.008\\
Hero &0.72 &0.90& 0.93 &0.94 &0.95 $\pm$ 0.011&\textbf{0.96} $\pm$ 0.01\\
Montezuma Revenge & 0.41 &0.76& 0.78 &\textbf{0.84} &0.82 $\pm$ 0.01&\textbf{0.84} $\pm$ 0.008\\
Ms Pacman &0.60& 0.67 &0.73 &0.77 &0.76 $\pm$ 0.016&\textbf{0.78} $\pm$ 0.012\\
Pitfall &0.35& 0.49 &0.61 &\textbf{0.74} &0.72 $\pm$ 0.027&\textbf{0.74} $\pm$ 0.023\\
Pong&0.19& 0.73& 0.82&0.85 &\textbf{0.86} $\pm$ 0.005&0.85 $\pm$ 0.004\\
Private Eye &0.72& 0.81 &0.91&0.93 &0.92 $\pm$ 0.012&\textbf{0.94} $\pm$ 0.01\\
Qbert &0.53 &0.66 &0.74 &\textbf{0.8} &\textbf{0.8} $\pm$ 0.023 &\textbf{0.8} $\pm$ 0.02\\
Seaquest &0.61&0.69 &0.69 &0.7 &0.69 $\pm$ 0.009 &\textbf{0.72} $\pm$ 0.009\\
Space Invaders &0.57 &0.57 &0.59&0.63 &0.61 $\pm$ 0.02 &\textbf{0.7} $\pm$ 0.013\\
Tennis &0.37 &0.61 &0.61 &0.65 &0.65 $\pm$ 0.007 &\textbf{0.72} $\pm$ 0.008\\
Venture & 0.43 & 0.52 & 0.59 &0.59 &0.6 $\pm$ 0.02 &\textbf{0.62} $\pm$ 0.024\\
Video Pinball & 0.47 & 0.59 & 0.61 &0.63 &\textbf{0.68} $\pm$ 0.02&0.67 $\pm$ 0.018\\\hline
Mean & 0.46 & 0.65 & 0.73 & 0.76 & 0.76 $\pm$ 0.015 &\textbf{0.79} $\pm$ 0.014\\
\hline
\end{tabular}
\label{tbl:reso}
\end{table*}

Table \ref{tbl:reso} presents supplementary results on accuracy scores from linear evaluations across each game, which are similar to the F1 score trends observed in Table \ref{tbl:resf1o}. The results for VAE, CPC, and ST-DIM are taken from \cite{anand2019unsupervised}, and the results for DIM-UA are taken from \cite{meng2023state}. In Table \ref{tbl:reso}, DIM-C$^+$ demonstrates consistent performance, matching or surpassing the results of all other assessed methods in 15 out of 19 games and exceeding the benchmark accuracy set by DIM-UA in 13 games. The average accuracy score of DIM-C$^+$ is 79\%, also the highest among all methods, and higher than the 76\% mark established by the previous state-of-the-art method DIM-UA.

The accuracy scores in Table \ref{tbl:reso} together with the F1 scores in Table \ref{tbl:resf1o} both underscore the substantial improvements by exploiting a capacity regularizer alone. Meanwhile, DIM-UAC$^+$ gains equal or higher accuracy scores than DIM-UA in 10 out of 19 games, but only higher accuracy scores than DIM-UA in seven games. Both DIM-UA and DIM-UAC$^+$ have an identical mean accuracy score pinned at 76\%, indicating the limited improvement through combining C$^+$ with UA as well.

\end{document}